\documentclass[journal]{IEEEtran}

\usepackage{graphicx}

\usepackage{tikz}
\usepackage{comment}
\usepackage{booktabs}
\usepackage{multirow}
\usepackage{amsmath,amssymb} 
\usepackage[linesnumbered,ruled,vlined]{algorithm2e}
\usepackage{color}
\usepackage[colorlinks,bookmarks=false,citecolor=green]{hyperref}

\newcommand{\eg}{\emph{e.g.}}
\newcommand{\ie}{\emph{i.e.}}

\usepackage[normalem]{ulem}

\begin{document}
%
\title{DMMG: Dual Min-Max Games for Self-Supervised Skeleton-Based Action Recognition}



%
\author{\IEEEauthorblockN{Shannan GUAN\IEEEauthorrefmark{1},
Xin YU\IEEEauthorrefmark{2},
Wei HUANG\IEEEauthorrefmark{3}, 
Gengfa FANG\IEEEauthorrefmark{4}, and
Haiyan LU\IEEEauthorrefmark{1}}\\
\IEEEauthorblockA{\IEEEauthorrefmark{1} Australia Artificial Intelligence Institute\\
University of Technology Sydney, Australia, AU}
\IEEEauthorblockA{\IEEEauthorrefmark{2}School of Information Technology and Electrical Engineering\\
University of Queensland, Australia, AU}
\IEEEauthorblockA{\IEEEauthorrefmark{3}RIKEN Center for Advanced Intelligence Project, Tokyo, Japan, JP}
\IEEEauthorblockA{\IEEEauthorrefmark{4}School of Electrical and Data Engineering\\
University of Technology Sydney, Australia, AU}}


\maketitle



%
\IEEEpeerreviewmaketitle

\begin{abstract}
In this work, we propose a new Dual Min-Max Games (DMMG) based self-supervised skeleton action recognition method by augmenting unlabeled data in a contrastive learning framework.
Our DMMG consists of a viewpoint variation min-max game and an edge perturbation min-max game. 
These two min-max games adopt an adversarial paradigm to perform data augmentation on the skeleton sequences and graph-structured body joints, respectively. 
Our viewpoint variation min-max game focuses on constructing various hard contrastive pairs by generating skeleton sequences from various viewpoints. 
These hard contrastive pairs help our model learn representative action features, thus facilitating model transfer to downstream tasks.
Moreover, our edge perturbation min-max game specializes in building diverse hard contrastive samples through perturbing connectivity strength among graph-based body joints.
The connectivity-strength varying contrastive pairs enable the model to capture minimal sufficient information of different actions, such as representative gestures for an action while preventing the model from overfitting.
By fully exploiting the proposed DMMG, we can generate sufficient challenging contrastive pairs and thus achieve discriminative action feature representations from unlabeled skeleton data in a self-supervised manner.
Extensive experiments demonstrate that our method achieves superior results under various evaluation protocols on widely-used NTU-RGB+D and NTU120-RGB+D datasets.
\end{abstract}

\section{Introduction}

\begin{figure}[t]
    \centering
    \includegraphics[width=1\linewidth]{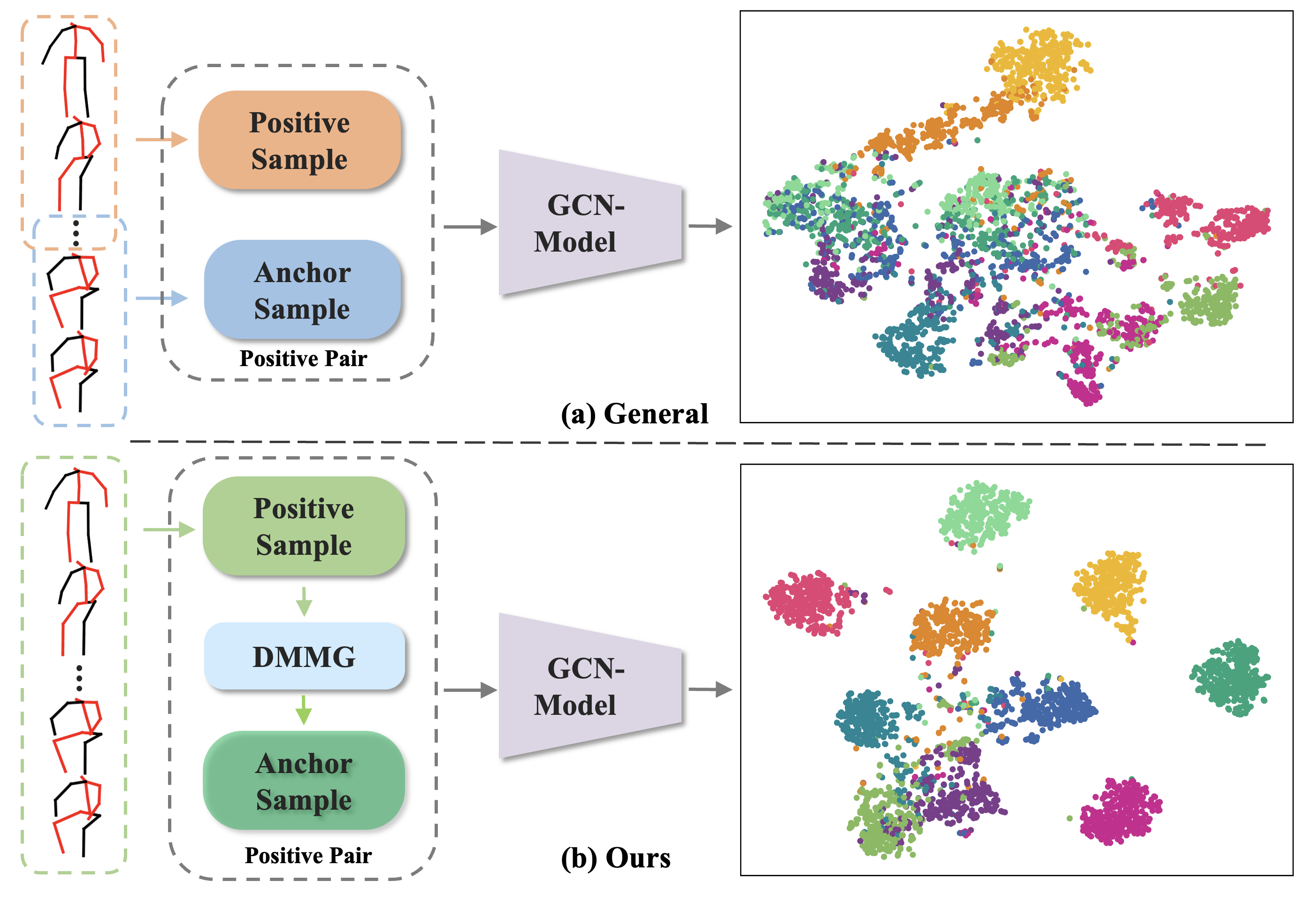}
    \caption{Comparison of conventional skeleton data augmentation and our DMMG. (a) the generic skeleton-based data augmentation (\eg, Crop or Shear) and the learned feature distribution; (b) our proposed data augmentation method DMMG and the distribution of our features.}
    \label{fig:introcompare}
\end{figure}

Skeleton-based human action recognition has attracted many researchers' attention for decades~\cite{survey2}.
With the advance of depth cameras (e.g., Kinect sensor) and robust 3D pose estimation methods~\cite{openpose,3dpose1,simple3dpose}, 3D skeleton data becomes more accessible. 
In the past few years, many fully-supervised skeleton-based action recognition methods~\cite{asgcn,actrec1,actrec2,actrec3,actrec4,actrec5,guan2022256} achieve high accuracy by elaborately designing models. 
However, fully-supervised methods heavily rely on a large amount of annotated skeleton data, but annotating data is high-cost and time-consuming. 
To complement the lack of annotated skeleton data, some recent action recognition works~\cite{unsupact2,unsupact3,unsupact5,extremeaug,crossviews,skeletonmix} have delved into exploring self-supervised approaches. 

In the self-supervised skeleton-based action recognition task, contrastive learning~\cite{conl1,contheory} has been introduced to learn feature representations. 
Those self-supervised action recognition methods~\cite{extremeaug,crossviews,mccclr,skeletonmix} first augment skeleton data to construct contrastive pairs (\ie, positive and negative pairs). 
Then, they learn action features by increasing the similarity between positive pairs while decreasing the similarity between negative pairs using a contrastive loss~\cite{mocoloss,infonce,onlinemining}. 

Although those self-supervised action recognition methods achieve promising results, 
they still face the following challenges: 
(1) Their data augmentation strategies are based on the random mechanism. 
Those methods may not generate sufficient hard contrastive pairs to learn representative action features~\cite{gcnaug1,gcnaug2,gcnaug3}. 
(2) Existing skeleton data augmentation strategies~\cite{extremeaug,crossviews,mccclr} may construct misleading positive pairs, and these pairs can cause ambiguity in model learning.
For example, the \emph{crop} operation may slice the sit-down action into two parts: a sit sequence and a stand sequence, and using these two sequences as positive pairs would harm model learning.
(3) Several methods may not focus on the most discriminative and representative features~\cite{infomax}.
For example, many people habitually swing their arms while walking, but learning swinging arms may not best represent walking actions. How to effectively capture the most representative motions for action recognition still remains challenging.

In this paper, we propose a viewpoint variation min-max game and an edge perturbation min-max game, and we combine them as Dual Min-Max Games (DMMG) to address the aforementioned three challenges.
Our two min-max games are built on an adversarial paradigm to perform data augmentation on the skeleton sequences and graph-structured body joints, respectively. 
We can construct a large number of sufficient challenging contrastive pairs by fully exploiting our DMMG, and we leverage these hard contrastive pairs to learn more representative and discriminative action feature representations. 
In this fashion, we can improve the model performance on downstream tasks.

To be specific, our viewpoint variation min-max game augments skeleton sequences from various viewpoints. 
We first try to find various viewpoints to increase the visual differences for an identical 3D skeleton sequence.
Then, we maximize the similarity between the feature representations of those skeleton sequences under different viewpoints.
As our viewpoint variation min-max game only finds a new viewpoint to render original skeletons, we can preserve the structural and temporal consistency of the skeleton sequences during the data augmentation.
In this way, our constructed positive pairs do not have ambiguity and they will facilitate learning distinctive action features.

Our edge perturbation min-max game augments graph-based body joints to construct connectivity-strength varying contrastive pairs by perturbing their connectivity strengths. 
It minimizes the correspondence between the original and augmented graph-based body joints by reducing the connectivity strengths and then maximizes the similarity between their feature representations~\cite{ib1,ib2,ib3}. 
For example, we can augment graph-based body joints of a walking action with minimal connectivity strengths among arm-related joints while preserving the connectivity strengths among the leg-related joints.
In this manner, important joint connections will be highlighted to represent an action. 
Therefore, our edge perturbation min-max game will learn to identify representative joint connections for action recognition. 
Moreover, thanks to these connectivity-strength varying contrastive pairs, we can better prevent overfitting.

As shown in Fig.~\ref{fig:introcompare}, the action features represented by our DMMG are more discriminative than other methods.
Moreover, extensive experimental results on NTU-RGB+D datasets~\cite{NTURGB60,NTURGB120} demonstrate that our DMMG boosts the model's performance on various downstream tasks. 
Our main contributions can be summarised as follows: 
\begin{itemize}
\item We propose a novel Dual Min-Max Games (DMMG) under an adversarial paradigm for self-supervised skeleton action recognition in a contrastive learning framework. 
\item Our DMMG performs a challenging data augmentation on the skeleton sequences and graph-structured body joints. It constructs a large number of unambiguous hard contrastive pairs in learning discriminative action features. 
\item Our DMMG constructs connectivity-strength varying contrastive pairs and allows us to capture pivotal representations of actions while avoiding overfitting issues.
\item Our DMMG achieves state-of-the-art performance under various evaluation protocols on two benchmark datasets: NTU RGB+D 60 and NTU RGB+D 120.
\end{itemize}
\section{Related Work}
\subsection{Self-Supervised Skeleton-Based Action Recognition}
Self-supervised skeleton-based action recognition aims to learn action feature representations from numerous unlabeled 3D skeleton sequences. 
Due to the lack of label supervision, self-supervised learning generally requires an intra-supervision signal derived from unlabeled data~\cite{crossviews}.
In the early stage, some methods generate supervision signals by designing pretext tasks, such as reconstructing skeleton sequence~\cite{unsupact2}, using colorization~\cite{colorization}, autoregression~\cite{unsupact3,pandc}, prediction motion~\cite{unsupact6}, and jigsaw puzzles~\cite{unsupact1,unsupact4}. 
To explore more useful supervision signals, some methods reconstruct skeleton sequences by using a generative adversarial network (GAN). 
For example, LongT GAN~\cite{longtgan} proposes an auto-encoder-based GAN to reconstruct the sequential information of skeletons. 
P\&C~\cite{pandc} utilizes an encoder-decoder framework to reconstruct a skeleton sequence and learn action features.
Colorization~\cite{colorization} designs a colorized point cloud to represent skeleton data and utilizes an auto-encoder framework to learn spatio-temporal features from these hand-crafted colorized skeleton joints.
MS$^2$L~\cite{ms2l} develops a multi-task learning framework by combining contrastive learning with pretext tasks. 
However, these methods often rely on pretext tasks and may not generalize well on downstream tasks.

Recently, the contrastive learning mechanism has demonstrated promising performance in self-supervised action recognition.
The supervision signals of the contrastive learning paradigm are usually generated by a contrastive loss~\cite{contheory}, such as InfoNCE~\cite{infonce}, SimCLR~\cite{simclr}, MoCo~\cite{mocoloss}, and OTM~\cite{onlinemining}. 
These contrastive losses have been widely used in recent self-supervised action recognition methods.
For example, CrosSCLR~\cite{crossviews} and SkeletonMixCLR~\cite{skeletonmix} achieve promising performance by adopting MoCov2~\cite{mocov2}. 
AimCLR~\cite{extremeaug} adopts InfoNCE~\cite{infonce} as a contrastive loss to improve the model performance on downstream tasks.
In this paper, we use OTM~\cite{onlinemining} to implement our DMMG.

\subsection{Skeleton Data Augmentation Strategy}
Constructing contrastive pairs via skeleton data augmentation is a critical component in self-supervised action recognition tasks under a contrastive learning framework. 
Most contrastive learning-based self-supervised action recognition methods focus on developing various data augmentation strategies to construct hard contrastive pairs, and the harder contrastive pairs would promote learning more representative action features.
For example, MCC~\cite{mccclr} adopts a speed-changed operation combined with a random start frame to build hard positive samples. 
CrosSCLR~\cite{crossviews} adopts \emph{Shear} and \emph{Crop} data augmentation, which has become the most commonly used skeleton data augmentation strategies in self-supervised action recognition works~\cite{crossviews,skeletonmix,extremeaug}. 

To explore more discriminative action feature representations. 
AimCLR~\cite{extremeaug} adopts more hand-crafted data augmentation approaches, such as \emph{Spatial Flipping, Rotation, Axis Masking, Temporal Flipping}, to construct hard contrastive pairs. 
Furthermore, SkeletonMixCLR~\cite{skeletonmix} develops a stronger skeleton data augmentation method by randomly mixing different body parts (e.g., left hands, right legs, trunk) among different skeleton sequences. 
Although these hand-crafted data augmentation methods can create effective contrastive pairs, their random data augmentation strategies may construct sufficient challenging contrastive pairs. 
On the contrary, our DMMG adopts an adversarial learning technique to optimize the data augmentation strategy and thus can construct a large number of non-misleading yet challenging contrastive pairs. 
\section{Preliminary}
In this section, we introduce the preliminary concepts and definitions of our DMMG. 
Here, we denote a 3D skeleton sequence as $\mathbf{X} \in \mathbb{R}^{J\times C\times T} $, which contains $T$ frames with $J$ joints, and each joint has $C=3$ dimensions, \emph{i.e.}, $(x,y,z)$ coordinates. 
Then, we represent a graph structure $\mathcal{G}=(\mathcal{V}, \mathcal{E})$, where $v_i \in \mathcal{V}$ is a node and $\left(v_i, v_j\right) \in \mathcal{E}$ is an edge. 
The node features are 3D coordinates of skeleton joints, and we use an adjacency matrix to represent graph-structured body joints. 
The adjacency matrix is denoted by $\mathbf{A} \in \mathbb{R}^{J \times J}$, whose element $\mathbf{A}_{i j}$ associates with the edge $\left(v_i, v_j\right)$.
The conventional GCN-based models generally only adopt the node features as input, and the adjacency matrix is applied to hidden layers. 
To implement our DMMG, we modify the GCN-based model which can take both node features and adjacency matrix as inputs, formulated as $f(\mathbf{X},\mathbf{A})$. 
We aim to learn a model $f: \mathcal{G} \rightarrow \mathbb{R}^d$ for further use in downstream tasks.
\subsection{Viewpoint Variation Min-Max Game}
In this viewpoint variation min-max game, we first define a skeleton augmenter $\mathcal{R}$.
$\mathcal{R}$ takes $\mathbf{X}$ as inputs and generates $\widetilde{\mathbf{X}} = \mathcal{R} \left(\mathbf{X} \right) \in \mathbb{R}^{J\times C\times T}$.
The minimization target is to minimize the mutual information~\cite{adgcl} between the feature representations $f(\mathbf{X},\mathbf{A})$ and $f(\widetilde{\mathbf{X}},\mathbf{A})$ by training the skeleton augmenter $\mathcal{R}$. 
It can be formulated as: 
\begin{equation}
\label{rotmingame}
\min_{\mathcal{R}} I(f(\mathbf{X},\mathbf{A});f(\mathcal{R} \left(\mathbf{X} \right),\mathbf{A})),
\end{equation}
where $I(\cdot~; \cdot)$ denotes the mutual information between two feature representations.

This maximization game is to maximize the mutual information between the feature representations $f(\mathbf{X},\mathbf{A})$ and $f(\widetilde{\mathbf{X}},\mathbf{A})$ by training the model $f$, and it can be expressed as follows:
\begin{equation}
\label{rotmaxgame}
\max_f I(f(\mathbf{X},\mathbf{A});f(\mathcal{R} \left(\mathbf{X} \right),\mathbf{A})).
\end{equation}
Combining Eq.~\eqref{rotmingame} and Eq.~\eqref{rotmaxgame}, our viewpoint variation min-max game is formulated as:
\begin{equation}
\min_{\mathcal{R}}\max_f I(f(\mathbf{X},\mathbf{A});f(\mathcal{R} \left(\mathbf{X} \right),\mathbf{A})).
\end{equation}

\subsection{Edge Perturbation Min-Max Game}
In the edge perturbation min-max game, we first present a learnable graph augmenter $\mathcal{T}$ with inputs $\mathbf{A}$ and $\mathbf{X}$. 
For an adjacency matrix $\mathbf{A}$, $\widetilde{\mathbf{A}}=\mathcal{T}(\mathbf{A},\mathbf{X})$ denotes the perturbed adjacency matrix, where $\widetilde{\mathbf{A}}_{i j}$ remains the same connection but has different weights for the edge $\left(v_i, v_j\right)$. 
Here, the minimization game aims to minimize the mutual information between the feature representations $f(\mathbf{X},\mathbf{A})$ and $f(\mathbf{X},\mathcal{T}(\mathbf{A},\mathbf{X}))$ by training the graph augmenter $\mathcal{T}$, defined by: 
\begin{equation}
\label{edgemingame}
\min_{\mathcal{T}} I(f(\mathbf{X},\mathbf{A});f(\mathbf{X},\mathcal{T}(\mathbf{A},\mathbf{X}))).
\end{equation}

The maximization game is to maximize the mutual information between the feature representations $f(\mathbf{X},\mathbf{A})$ and $f(\mathbf{X},\mathcal{T}(\mathbf{A},\mathbf{X}))$ by training the model $f$, formulated as:
\begin{equation}
\label{edgemaxgame}
\max_f I(f(\mathbf{X},\mathbf{A});f(\mathbf{X},\mathcal{T}(\mathbf{A},\mathbf{X}))).
\end{equation}
Combining Eq.~\eqref{edgemingame} and Eq.~\eqref{edgemaxgame}, our edge perturbation min-max game is defined as:
\begin{equation}
\min_{\mathcal{T}}\max_f I(f(\mathbf{X},\mathbf{A});f(\mathbf{X},\mathcal{T}(\mathbf{A},\mathbf{X}))).
\end{equation}

\section{Methodology}
\begin{figure}[t]
    \centering
    \includegraphics[width=0.9\linewidth]{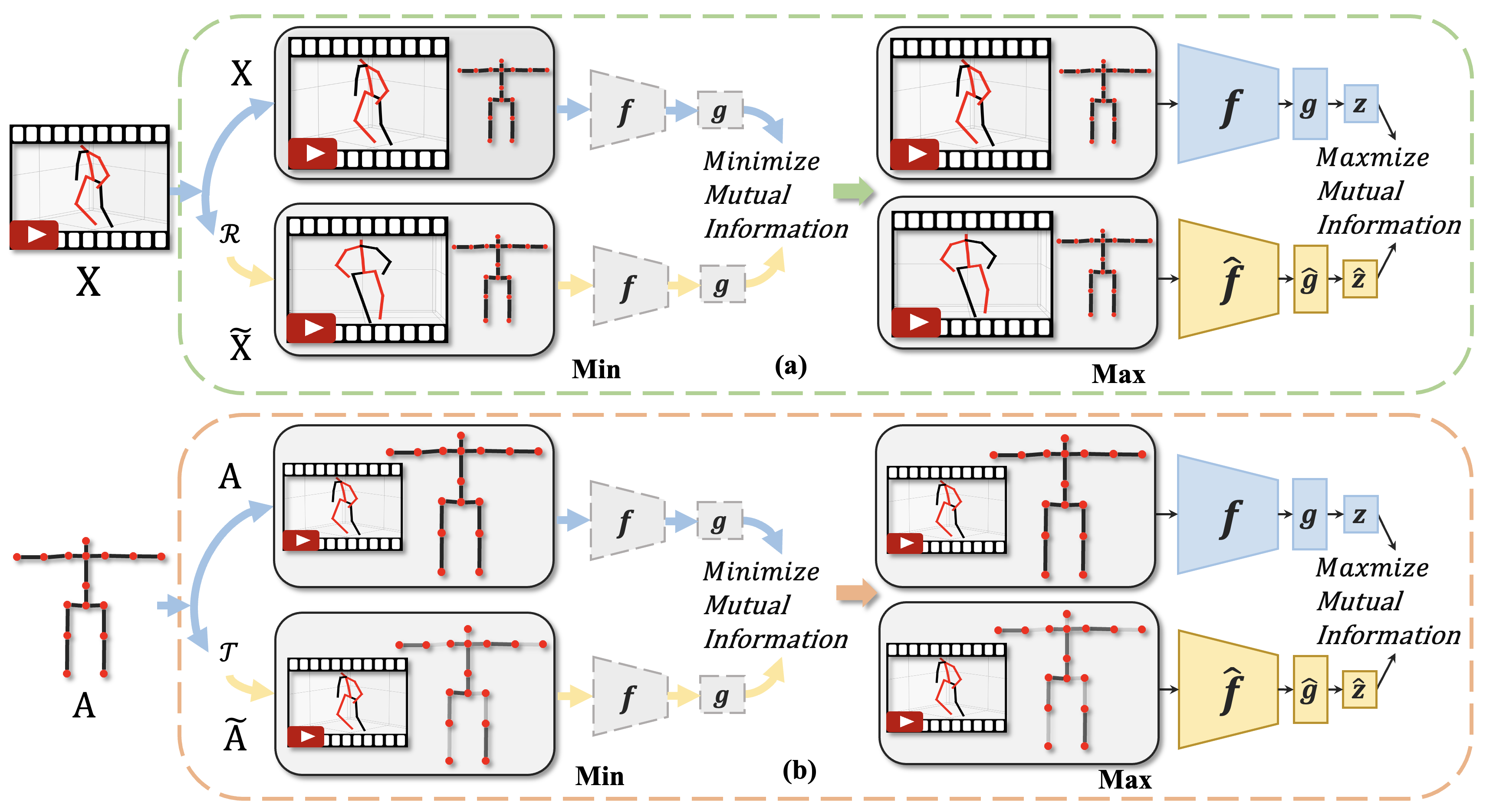}
    \caption{Pipeline of our Dual Min-Max Games (DMMG). Left: original inputs; Right: 
    (a) block illustrates the viewpoint variation min-max game, and (b) block shows the edge perturbation min-max game. The models with gray color denote their parameters are fixed. }
    \label{fig:pipeline}
\end{figure}

In this section, we first introduce the data streams and model used in our DMMG, and then present the details of our two min-max games (e.g., the details of key modules, principles and algorithms) as well as the contrastive learning framework to implement our DMMG. 

\subsection{Model Details}
\begin{figure*}[t]
    \centering
    \includegraphics[width=0.85\linewidth]{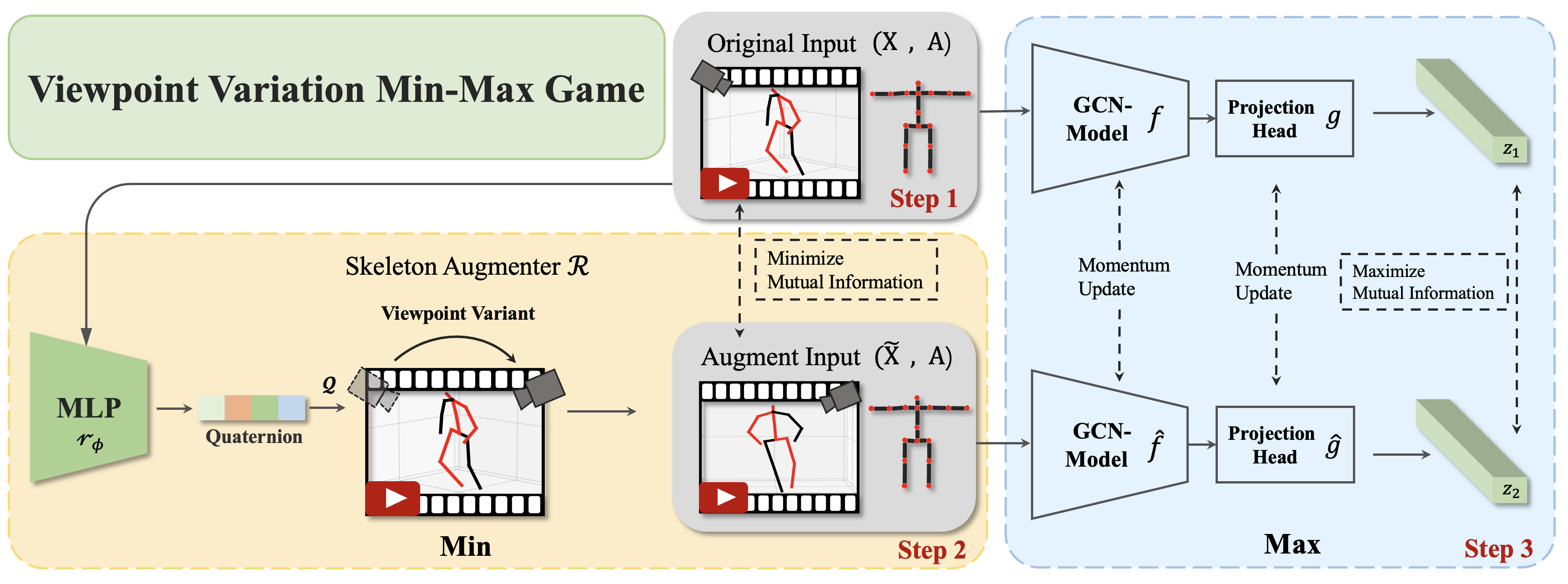}
    \caption{Illustration of our viewpoint variation min-max Game. In step 1, we apply the minimization game to the original skeleton data. Then, the minimization game guides the skeleton augmenter $\mathcal{R}$ to minimize the mutual information between them in step 2. Finally, the maximization game leads the GCN model to maximize the mutual information between the original skeleton data and augmented skeleton data in step 3. }
    \label{fig:viewvariant}
\end{figure*}

\subsubsection{Data Stream of DMMG}
Skeleton-based data can be easily converted into various data streams~\cite{twostreamagcn}. 
We utilize the joint and motion streams as inputs.
The joint stream is noted as $\mathbf{X}=\left\{\mathbf{X}_1, \mathbf{X}_2, \ldots, \mathbf{X}_T\right\}$ and the motion stream is represented as the joint displacement between frames: $\mathbf{X}_{t}-\mathbf{X}_{t-1}$. 
For constructing contrastive pairs, we consider $\left\{(\mathbf{X},\mathbf{A});(\widetilde{\mathbf{X}},\mathbf{A})\right\}$ as a positive pair in our viewpoint variation min-max game, and any other samples are regarded as negative pairs. 
In our edge perturbation min-max game, we define $\left\{ (\mathbf{X},\mathbf{A});(\mathbf{X},\widetilde{\mathbf{A}})\right\}$ as a positive pair, and any other samples serve as negative pairs.

\subsubsection{Model $f$}
In this work, we employ ST-GCN~\cite{stgcn} as the model.
ST-GCN can explore the spatio-temporal information in 3D skeleton sequences, and it is widely used in self-supervised 3D skeleton action recognition methods~\cite{crossviews,skeletonmix,extremeaug}. 
The GCN based model $f$ embeds $\mathbf{X}$ with the pre-defined adjacency matrix $\mathbf{A}$ into a hidden space $\mathbf{H} = f(\mathbf{X},\mathbf{A})$.
In particular, the GCNs can be represented with a simple form~\cite{gcn}:
\begin{equation}
\mathbf{H}^{(l+1)}=f\left(\mathbf{H}^{(l)}, \mathbf{A}\right)=\sigma\left(\mathbf{\Lambda}^{-\frac{1}{2}} \mathbf{A} \mathbf{\Lambda}^{-\frac{1}{2}}\mathbf{H}^{(l)} \mathbf{W}^{(l)}\right),
\label{gcns}
\end{equation}
where $\mathbf{\Lambda}_{i i}=\sum_j\mathbf{A}_{i j}$;
$\mathbf{H}^{(l)}$ is the output feature from the $l$-th GCN layer, when $l=0$ and $\mathbf{H}^{(0)}= \mathbf{X} \in \mathbb{R}^{{J\times C\times T}}$; 
$\sigma\left(\cdot \right)$ denotes a nonlinear function, i.e., ReLu function;
$\mathbf{W}^{(l)}$ is the weights matrix in the $l$-th GCN layer.  
We can substitute the original $\mathbf{A}$ with augmented $\widetilde{\mathbf{A}}$, and \eqref{gcns} can be modified as:
\begin{equation}
\mathbf{H}^{(l+1)}=f\left(\mathbf{H}^{(l)}, \widetilde{\mathbf{A}}\right)=\sigma\left(\mathbf{\Lambda}^{-\frac{1}{2}} \widetilde{\mathbf{A}} \mathbf{\Lambda}^{-\frac{1}{2}}\mathbf{H}^{(l)} \mathbf{W}^{(l)}\right).
\end{equation}

The output features can be extracted by several GCN layers, represented as $\boldsymbol{z}_{out} \in \mathbb{R}^{C_{out} \times T_{out} \times V}$.  
Then, the GCN model is followed by an average pooling operation on the spatio-temporal dimension, and the action feature is finally formulated as $z\in \mathbb{R}^d$.

\begin{figure*}[t]
    \centering
    \includegraphics[width=0.85\linewidth]{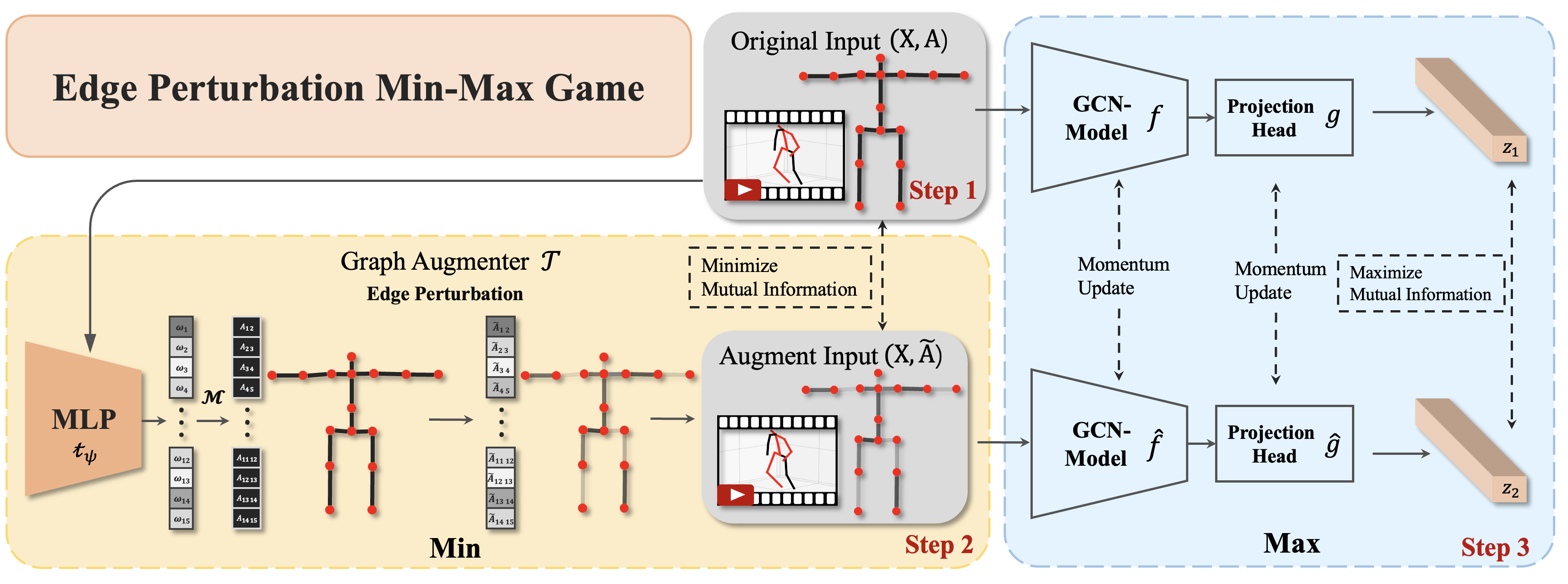}
    \caption{Illustration of the Edge Perturbation Min-Max Game. In step 1, we apply the minimization game to the skeleton data and original graph-structured body joints. Then the minimization game leads the graph augmenter $\mathcal{T}$ to minimize the mutual information between them. Finally, the maximization game guides the model to maximize the mutual information between the original and augmented graph-structured body joints in step 3.}
    \label{fig:edgeperturbation}
\end{figure*}

\IncMargin{1em}
\begin{algorithm}[t]  
\small
    \SetKwInOut{Input}{input}\SetKwInOut{Output}{output}
	
	\Input{Data $\mathbf{X} \in \mathbb{R}^{J \times C \times T}$, $\mathbf{A} \in \mathbb{R}^{J \times J}$; Model $f_{\theta}$, Projector $g_{\omega}$, Skeleton Augmenter $\mathcal{R}$,
    Mutual Information $I$, Hyper-Parameters $\gamma$
 }
      \Output{Trained Model $f_{\theta}$}
      \Begin{
	 	\For{sampled minibatch $\mathcal{G}_{n}=(\mathbf{X}_{n},\mathbf{A}_{n}): n=1,2 \ldots N$}{
            \emph{/* viewpoint variation minimization game */}\\
            \For{$n=1$ \KwTo $N$}{ 
            $\widetilde{\mathbf{X}}_{n}$ = $\mathcal{R}(\mathbf{X}_{n})$ \\
            $z_{n, 1},z_{n, 2}=g(f(\mathbf{X}_{n},\mathbf{A}_{n})),g(f(\widetilde{\mathbf{X}}_{n},\mathbf{A}_{n}))$ \\
            }
            $z_{1}$ , $z_{2}$ = $\{z_{n, 1}\}^{N}$ , $\{z_{n, 2}\}^{N}$\\
            $\mathcal{L}_{min}$ = $I(z_{1},z_{2})$ \\
            update augmenter parameters via gradient ascent \\
            \emph{/* viewpoint variation maximization game */}\\
            \For{$n=1$ \KwTo $N$}{ 
            $\widetilde{\mathbf{X}}_{n}$ = $\mathcal{R}(\mathbf{X}_{n})$ \\
            $z_{n, 1},z_{n, 2}=g_{\omega}(f_{\theta}(\mathbf{X}_{n},\mathbf{A}_{n})),\hat{g}_{\hat{\omega}}(\hat{f}_{\hat{\theta}}(\widetilde{\mathbf{X}}_{n},\mathbf{A}_{n}))$ \\
            }
            $z_{1}$ , $z_{2}$ = $\{z_{n, 1}\}^{N}$ , $\{z_{n, 2}\}^{N}$\\
            $\mathcal{L}_{max}$ = $\gamma$ - $I(z_{1},z_{2})$ \\
            update model $f_{\theta}$,$\hat{f}_{\hat{\theta}}$, and projection head $g_{\omega}$,$\hat{g}_{\hat{\omega}}$ parameters via gradient descent
   
   \Return{Model $f_{\theta}$}
}}
    \caption{Viewpoint Variation Min-Max Game}
    \label{algo_vpvmm} 
\end{algorithm}
\DecMargin{1em}

\subsection{Dual Min-Max Games}

\subsubsection{Viewpoint Variation Min-Max Game}

As shown in Fig.~\ref{fig:viewvariant}, in the skeleton augmenter $\mathcal{R}$, a skeleton sequence $\mathbf{X}$ is firstly fed into a multi-layer perceptron (MLP): $r_{\phi}$ with trainable parameters $\phi$, and outputs a normalized quaternion $\boldsymbol{q} \in \mathbb{R}^{1 \times 4}$~\cite{quaternion}. 
Then, $\boldsymbol{q}$ is used for rotating the skeleton data $\mathbf{X}$ and generating the augmented skeleton data $\widetilde{\mathbf{X}}$. 
This quaternion rotation operation has the same effect as changing the viewpoints. 
We denote such a quaternion rotation operation as $\mathcal{Q}$ and the augmented skeleton sequence is generated by $\widetilde{\mathbf{X}} = \mathcal{Q}(\mathbf{X},\boldsymbol{q})$. 
Note that in the viewpoint variation min-max game, we use the original adjacency matrix for the original and augmented skeleton data to construct the contrastive pair: $\left\{(\mathbf{X},\mathbf{A});(\widetilde{\mathbf{X}},\mathbf{A})\right\}$. 
They are used as inputs for GCN model $f_{\theta}$ to learn action features.
In the minimization game, we train $r_{\phi}$ for generating harder contrastive pairs by minimizing the mutual information between the original and augmented action features. Noted that the parameter $\theta$ of GCN model $f_{\theta}$ are fixed. Eq.~\eqref{rotmingame} can be modified as: 
\begin{equation}
\label{rotmingame-2}
\min_{\phi} I(f_{\theta}(\mathbf{X},\mathbf{A});f_{\theta}(\mathcal{Q}(\mathbf{X},r_{\phi}(\mathbf{X})),\mathbf{A})).
\end{equation}
 
In the maximization game, we train the GCN model $f_{\theta}$ to maximize the mutual information between the hard contrastive pairs. 
The parameter of $r_{\phi}$ is fixed in the maximization game, and Eq.~\eqref{rotmaxgame} is revised as: 
\begin{equation}
\label{rotmaxgame-2}
\max_{\theta} I(f_{\theta}(\mathbf{X},\mathbf{A});f_{\theta}(\mathcal{Q}(\mathbf{X},r_{\phi}(\mathbf{X})),\mathbf{A})).
\end{equation}

Combining Eq.~\eqref{rotmingame-2} with Eq.~\eqref{rotmaxgame-2}, our viewpoint variation min-max game is formally represented as: 
\begin{equation}
\label{rotmmgame}
\min_{\phi}\max_{\theta} I(f_{\theta}(\mathbf{X},\mathbf{A});f_{\theta}(\mathcal{Q}(\mathbf{X},r_{\phi}(\mathbf{X})),\mathbf{A})).
\end{equation}

\subsubsection{Edge Perturbation Min-Max Game}

As shown in Fig.~\ref{fig:edgeperturbation}, the graph augmenter $\mathcal{T}$ takes $\mathbf{X}$ and $\mathbf{A}$ as the inputs. 
Firstly, $\mathbf{X}$ is fed into an MLP: $t_{\psi}$ with trainable parameters $\psi$, and outputs a vector $\boldsymbol{w} \in \mathbb{R}^{n}$, where $n$ is the number of pre-defined edges in the adjacency matrix and $\boldsymbol{w}$ is restricted in the range from 0 to 1 by a Sigmoid activation function. 
Then, $\boldsymbol{w}$ is multiplied with all non-zero elements in $\mathbf{A}$. 
Here, we denote this element-wise multiplication as $\mathcal{M}$, and thus we can obtain augmented graph-structure based body joints by $\widetilde{\mathbf{A}}=\mathcal{M}(\mathbf{A},\boldsymbol{w})$. 
In the minimization game, we train $t_{\psi}$ to generate more challenging contrastive pairs from the graph-structured data perspective. 
We minimize the mutual information between the original and augmented graph-structured body joints embedded by GCN model $f_{\theta}$, where the parameter ${\theta}$ of the model is fixed. 
We revised Eq.~\eqref{edgemingame} as:

\begin{equation}
\label{edgemingame-1}
\min_{\psi} I(f_{\theta}(\mathbf{X},\mathbf{A});f_{\theta}(\mathbf{X},\mathcal{M}(\mathbf{A},t_{\psi}(\mathbf{X})))).
\end{equation}

In the maximization game, we train the model $f_{\theta}$ to maximize the mutual information between hard contrastive pairs. Different from the minimization game, we fix the parameter ${\psi}$ of $t_{\psi}$ and Eq.~\eqref{edgemaxgame} is revised as:

\begin{equation}
\label{edgemaxgame-1}
\max_{\theta} I(f_{\theta}(\mathbf{X},\mathbf{A});f_{\theta}(\mathbf{X},\mathcal{M}(\mathbf{A},t_{\psi}(\mathbf{X})))).
\end{equation}

Combining Eq.~\eqref{edgemingame-1} and Eq.~\eqref{edgemaxgame-1}, we finally formulate our edge perturbation min-max game as:
\begin{equation}
\label{edgemmgame}
\min_{\psi} \max_{\theta} I(f_{\theta}(\mathbf{X},\mathbf{A});f_{\theta}(\mathbf{X},\mathcal{M}(\mathbf{A},t_{\psi}(\mathbf{X})))).
\end{equation}

\IncMargin{1em}
\begin{algorithm}[t]  
\small
    \SetKwInOut{Input}{input}\SetKwInOut{Output}{output}
	
	\Input{Data $\mathbf{X} \in \mathbb{R}^{J \times C \times T}$, $\mathbf{A} \in \mathbb{R}^{J \times J}$; Model $f_{\theta}$, Projector $g_{\omega}$, Graph Augmenter $\mathcal{T}$,
    Mutual Information $I$, Hyper-Parameters $\gamma$
 }
      \Output{Trained Model $f_{\theta}$}
      \Begin{
	 	\For{sampled minibatch $\mathcal{G}_{n}=(\mathbf{X}_{n},\mathbf{A}_{n}): n=1,2 \ldots N$}{
            \emph{/* edge perturbation minimization game */}\\
            \For{$n=1$ \KwTo $N$}{ 
            $\widetilde{\mathbf{A}}_{n}$ = $\mathcal{T}(\mathbf{X}_{n},\mathbf{A}_{n})$ \\
            $z_{n, 1},z_{n, 2}=g(f(\mathbf{X}_{n},\mathbf{A}_{n})),g(f(\mathbf{X}_{n},\widetilde{\mathbf{A}}_{n}))$ \\
            }
            $z_{1}$ , $z_{2}$ = $\{z_{n, 1}\}^{N}$ , $\{z_{n, 2}\}^{N}$\\
            $\mathcal{L}_{min}$ = $I(z_{1},z_{2})$ \\
            update augmenter parameters via gradient ascent \\
            \emph{/* edge perturbation maximization game */}\\
            \For{$n=1$ \KwTo $N$}{ 
            $\widetilde{\mathbf{A}}_{n}$ = $\mathcal{T}(\mathbf{X}_{n},\mathbf{A}_{n})$ \\
            $z_{n, 1},z_{n, 2}=g_{\omega}(f_{\theta}(\mathbf{X}_{n},\mathbf{A}_{n})),\hat{g}_{\omega}(\hat{f}_{\hat{\theta}}(\mathbf{X}_{n},\widetilde{\mathbf{A}}_{n}))$ \\
            }
            $z_{1}$ , $z_{2}$ = $\{z_{n, 1}\}^{N}$ , $\{z_{n, 2}\}^{N}$\\
            $\mathcal{L}_{max}$ = $\gamma$ - $I(z_{1},z_{2})$ \\
            update model $f_{\theta}$,$\hat{f}_{\hat{\theta}}$, and projection head $g_{\omega}$,$\hat{g}_{\hat{\omega}}$ parameters via gradient descent
   
   \Return{Model $f_{\theta}$}
}}
    \caption{Edge Perturbation Min-Max Game}
    \label{algo_epmm} 
\end{algorithm}
\DecMargin{1em} 

\subsection{Learning Framework of DMMG}
As shown in Fig.~\ref{fig:pipeline}, our minimization games first lead the data augmentation modules to generate harder contrastive samples by minimizing the mutual information between the learned features from the original and augmented data.
Then, in our maximization games, we utilize a contrastive loss to drive the model to learn feature representations by pulling the positive pairs closer and forcing the negative pairs away in embedding space. 
From the maximization games, as shown in Fig.~\ref{fig:viewvariant} and Fig.~\ref{fig:edgeperturbation}, the two models $f$ and $\hat{f}$ encode the original data and augmented data, where the parameter of $\hat{\theta}$ in $\hat{f}$ is momentum updated~\cite{mocoloss} of $f:\hat{\theta} \leftarrow \alpha \hat{\theta}+(1-\alpha) \theta$, and $\alpha$ is the momentum coefficient.
Then, a projector $g_{\omega}$ and its momentum updated version $\hat{g}_{\hat{\omega}}$ project the hidden spaces into a lower dimension space: $\boldsymbol{z}$ and $\hat{\boldsymbol{z}}$, where the projector is a fully connected layer with a ReLU activation function. 

In Eq.~\eqref{rotmmgame} and Eq.~\eqref{edgemmgame}, we utilize the term mutual information~\cite{adgcl} to evaluate the similarity between different features.  
To estimate the similarity, we adopt the online triplet mining loss~\cite{onlinetripletloss} as the estimator, which is frequently used in contrastive learning tasks~\cite{onlinetripletloss,tripletloss1,tripletloss2}. 
During the training process, we set a minibatch of $n$ samples, let $z_{i, 1}=g\left(f_{\theta}\left(G_i\right)\right)$ and $z_{i, 2}=\hat{g}(\hat{f}_{\hat{\theta}}(\widetilde{\mathcal{G}}_i))$, where $\mathcal{G}$ and $\widetilde{\mathcal{G}}$ denote the original and augmented data respectively. 
Here, we define the feature $z_{i, 1}$ learned by any original data as the anchor sample, the augmented feature $z_{i, 2}$ as the positive sample, and any other original feature $z_{i^{\prime}}$ as the negative sample.

\begin{equation}
\begin{split}
& I(f_{\theta}\left(\mathcal{G}_i\right);\hat{f}_{\hat{\theta}}(\widetilde{\mathcal{G}}_i)) = \\ 
& \sum_{i=1,i^{\prime} \neq i}^n\left[\left\|z_{i, 1}-z_{i^{\prime}, }\right\|_2^2 - \left\|z_{i, 1}-z_{i, 2}\right\|_2^2+\alpha\right]_{+},    
\end{split}
\end{equation}
where $\alpha$ is the margin between positive and negative pairs, and $[\cdot]_{+}:=\max (\cdot, 0)$ denotes the standard Hinge loss~\cite{onlinemining}. 
Generally, selecting the hard contrastive pairs to manipulate triplet loss significantly improves the model generalization on downstream tasks~\cite{onlinetripletloss}.
Therefore, during training, we select hard positive sample features ${z_i^p}$ and hard negative sample features ${z_i^n}$ from a minibatch by computing $\operatorname{argmax}_{z_i^p}\left\|z_{i, 1}-z_{i, 2}\right\|_2^2$ and $\operatorname{argmin}_{z_i^n}\left\|z_{i, 1}-z_{i^{\prime}, 1}\right\|_2^2$.
In the minimization games, we calculate the minimizing information loss by $\mathcal{L}_{min} = I(f_{\theta}\left(\mathcal{G}_i\right);\hat{f}_{\hat{\theta}}(\widetilde{\mathcal{G}}_i))$. 
In the maximization games, we define the maximizing information loss as $\mathcal{L}_{max} = \gamma - I(f_{\theta}\left(\mathcal{G}_i\right);\hat{f}_{\hat{\theta}}(\widetilde{\mathcal{G}}_i))$, where $\gamma$ is the hyperparameter for regularizing the loss value.

In the training stage, we implement our two min-max games alternatively in the same minibatch. 
For example, we first implement the viewpoint variation min-max game and then the edge perturbation min-max game. 

\section{Experiments}
\begin{table*}[t]
    \caption{Linear evaluation results compared with various methods on NTU-60 and NTU-120 datasets from different data streams. \textbf{J}, \textbf{M}, and \textbf{B} represent joint, motion, and bone stream respectively. ``*s'' denotes the multi-streams fusion. $\Delta$ is the performance gain compared to the methods using the same stream data, and $\ddagger$ indicates the model is pre-trained on NTU-61-120.}
    \centering
    \scriptsize
    \tabcolsep1.7mm
    \begin{tabular}{l|c|ccc|ccc|cc|cc}
        \toprule
        \multirow{3}{*}{Method} & \multirow{3}{*}{Stream} & \multicolumn{6}{c|}{NTU-60} & \multicolumn{4}{c}{NTU-120} \\ \cline{3-12}
        &
        & \multicolumn{3}{c|}{Xsub} & \multicolumn{3}{c|}{Xview} & \multicolumn{2}{c|}{Xsub} & \multicolumn{2}{c}{Xset}     \\ &    
        & Acc. (\%)   & $\Delta$ & $\Delta$$\ddagger$  & Acc. (\%)   & $\Delta$ & $\Delta$$\ddagger$  & Acc. (\%)  & $\Delta$  & Acc. (\%)   & $\Delta$ \\ \midrule

        SkeletonCLR~\cite{crossviews} &\textbf{J}& 68.3 & \textcolor{red}{+13.8} & \textcolor{red}{+10.9} & 76.4 & \textcolor{red}{+10.7} & \textcolor{red}{+10.2} & 56.8& \textcolor{red}{+12.8} & 55.9 &\textcolor{red}{+14.2} \\ 
        CrosSCLR~\cite{crossviews} &\textbf{J}& 72.9 & \textcolor{red}{+9.2} & \textcolor{red}{+6.3} & 79.9 & \textcolor{red}{+7.2} & \textcolor{red}{+6.7} & 61.3& \textcolor{red}{+8.3} & 60.5 &\textcolor{red}{+9.6} \\
        AimCLR~\cite{extremeaug}  &\textbf{J}& 74.3 & \textcolor{red}{+7.8} & \textcolor{red}{+4.9} & 79.7 & \textcolor{red}{+7.4} & \textcolor{red}{+6.9} & 63.4& \textcolor{red}{+6.2} & 63.4 &\textcolor{red}{+6.7} \\
        SkeleMixCLR~\cite{skeletonmix}  &\textbf{J}& 79.6 & \textcolor{red}{+2.5} & \textcolor{green}{-0.4}& 84.4 & \textcolor{red}{+2.7} & \textcolor{red}{+2.2}& 67.4& \textcolor{red}{+2.2} & 69.6 &\textcolor{red}{+0.5} \\
        DMMG$\ddagger$ (Ours)  & \textbf{J} &79.2&\textcolor{red}{+2.9} &- & 86.6 & \textcolor{red}{+0.5} &- & - & - & -& - \\
        \textbf{DMMG (Ours)}  & \textbf{J} &\textbf{82.1} & - & \textcolor{green}{-2.9} &\textbf{87.1}& - & \textcolor{green}{-0.5} &\textbf{69.6}& - &\textbf{70.1}& - \\ \midrule

        SkeletonCLR~\cite{crossviews} &\textbf{M}& 53.3 & \textcolor{red}{+23.4} & \textcolor{red}{+19.6} & 50.8 & \textcolor{red}{+30.2} & \textcolor{red}{+28.9} & 39.6& \textcolor{red}{+23.7} & 40.2 &\textcolor{red}{+21.9} \\ 
        CrosSCLR~\cite{crossviews} &\textbf{M}& 72.5 & \textcolor{red}{+4.2} & \textcolor{red}{+0.4} & 77.6 & \textcolor{red}{+3.4}  & \textcolor{red}{+2.1} & 59.4& \textcolor{red}{+3.9} & 59.2 &\textcolor{red}{+2.9} \\
        AimCLR~\cite{extremeaug}  &\textbf{M}& 66.8 & \textcolor{red}{+9.9} & \textcolor{red}{+6.1}& 70.6 & \textcolor{red}{+10.4}  & \textcolor{red}{+9.1} & 57.3& \textcolor{red}{+6.0} & 54.4 &\textcolor{red}{+7.7} \\
        SkeleMixCLR~\cite{skeletonmix}  &\textbf{M}& 70.3 & \textcolor{red}{+6.4} & \textcolor{red}{+2.6} & 76.1 & \textcolor{red}{+4.9}  & \textcolor{red}{+3.6}& 49.7 & \textcolor{red}{+13.6} & 53.8 &\textcolor{red}{+8.3} \\
        DMMG$\ddagger$ (Ours)  & \textbf{M} &72.9&\textcolor{red}{+3.8} &- &79.7& \textcolor{red}{+1.3} & - & -& -  &-& -  \\
        \textbf{DMMG (Ours)}  & \textbf{M} &\textbf{76.7}& - & \textcolor{green}{-3.8} &\textbf{81.0}& -  & \textcolor{green}{-1.3} &\textbf{63.3}& - &\textbf{62.1}& - \\ \midrule

        2s-SkeletonCLR~\cite{crossviews} &\textbf{J}+\textbf{M}& 75.0 & \textcolor{red}{+9.2} & \textcolor{red}{+7.5} & 79.8 & \textcolor{red}{+9.5} & \textcolor{red}{+9.2} & 60.7 & \textcolor{red}{+12.0} & 62.6 &\textcolor{red}{+9.8} \\ 
        2s-CrosSCLR~\cite{crossviews} &\textbf{J}+\textbf{M}& 77.8 & \textcolor{red}{+6.4}  & \textcolor{red}{+4.7} & 83.4 & \textcolor{red}{+5.9} & \textcolor{red}{+5.6} & 66.7 & \textcolor{red}{+6.0} & 65.1 &\textcolor{red}{+7.3} \\
        3s-AimCLR~\cite{extremeaug}  &\textbf{J}+\textbf{M}+\textbf{B}& 78.9 & \textcolor{red}{+5.3}  & \textcolor{red}{+3.6} & 83.8 & \textcolor{red}{+5.5} & \textcolor{red}{+5.2}& 68.2 & \textcolor{red}{+4.5} & 68.8 &\textcolor{red}{+3.6} \\
        3s-SkeleMixCLR~\cite{skeletonmix}  &\textbf{J}+\textbf{M}+\textbf{B}& 81.0 & \textcolor{red}{+3.2}  & \textcolor{red}{+1.5} & 85.6 & \textcolor{red}{+3.7} & \textcolor{red}{+3.4}& 69.1& \textcolor{red}{+3.6} & 69.9 &\textcolor{red}{+2.5}  \\

        2s-DMMG$\ddagger$ (Ours)  & \textbf{J+M} &82.5&\textcolor{red}{+1.7} &- & 89.0& \textcolor{red}{+0.3} &- & -& - & - & - \\
        \textbf{2s-DMMG (Ours)}  & \textbf{J}+\textbf{M} &\textbf{84.2}& -  & \textcolor{green}{-1.7} &\textbf{89.3}& - & \textcolor{green}{-0.3} &\textbf{72.7}& - &\textbf{72.4}& - \\ \bottomrule
    \end{tabular}
    \label{tab:linearevaul}
\end{table*}

\begin{table}[t]
    \caption{Linear evaluation results comparison on NTU-60 dataset. $\ddagger$ indicates the model is pre-trained on NTU-61-120. \emph{Single-stream} denotes the best results achieved using a single data stream, \emph{Multi-stream} represents the best results achieved by using multiple data streams.}
    \centering
    \scriptsize
    \begin{tabular}{l|c|cc}
        \toprule
        Method                   & Backbone         & Xsub (\%)       & Xset (\%)        \\      \midrule
        \multicolumn{4}{c}{\emph{Single-stream}}      \\ 
        LongT GAN~\cite{longtgan} &GRU          &52.1	&56.4	\\
        MS$^2$L~\cite{ms2l}      &GRU           & 52.6          & -           \\
        PCRP~\cite{PCRP}         &RNN        & 53.9          & 63.5           \\
        AS-CAL~\cite{AS-CAL}     &LSTM           & 58.5          & 64.8           \\
        P\&C~\cite{pandc}        &RNN           & 50.7	         & 76.3           \\
        CRRL~\cite{crrl}        &GRU          & 67.6	& 73.8	         \\
        ISC~\cite{isc}          &GCN         & 76.3	         & 85.2           \\
        SkeleMixCLR+~\cite{skeletonmix}    &GCN      & 80.7	         & 85.5            \\
        \textbf{DMMG (Ours)}  &GCN  & \textbf{82.1}        & \textbf{87.1}            \\
        \hline
        \multicolumn{4}{c}{\emph{Multi-stream}}  \\ 
        3s-CrosSCLR (LSTM)~\cite{crossviews}          &LSTM          & 62.8          & 69.2           \\
        3s-SkeletonCLR~\cite{crossviews}       &GCN          & 75.0          & 79.8           \\
        3s-Colorization~\cite{colorization}    &GCN            & 75.2          & 83.1           \\
        3s-CrosSCLR$\ddagger$~\cite{crossviews}          &GCN          & 72.8          & 80.7           \\
        3s-CrosSCLR~\cite{crossviews}          &GCN          & 77.8          & 83.4           \\
        3s-SkeleMixCLR+~\cite{skeletonmix}     &GCN           & 82.7          & 87.1           \\ 
        \textbf{2s-DMMG (Ours)}    &GCN    & \textbf{84.2}          &  \textbf{89.3}         \\ \bottomrule
    \end{tabular}
    \label{tab:ntu60-le}
\end{table}	

To evaluate the effectiveness of our DMMG, we implement experiments on two benchmark 3D skeleton-based action recognition datasets: NTU RGB+D 60~\cite{NTURGB60}, and NTU RGB+D 61-120~\cite{NTURGB120}, with a minibatch size of 32.
For optimization, we train the model on the PyTorch framework with an SGD optimizer with an initial learning rate of 0.1, weight decay at 0.0001, and the momentum is 0.9. 

\subsection{Datasets}
\begin{itemize}
\item \textbf{NTU RGB+D 60 (NTU-60)}~\cite{NTURGB60}: It is a large scale dataset for 3D skeleton based action recognition, which is recorded by Kinect V2 sensors and each skeleton graph is depicted by $J=25$ joints. 
In detail, this dataset contains 56,880 3D skeleton sequences from 40 different performers. 
These 3D skeleton sequences cover 60 daily actions, including single-person actions, human-objective, and human-human interactions. 
We evaluate our methods by two evaluation metrics: cross-subject (\textbf{Xsub}: skeleton sequences with 20 specific subject IDs are used for training and the remaining samples for testing) and cross-view (\textbf{Xview}: skeleton sequences from camera 2 and 3 for training while the other samples from camera 1 for testing).

\item \textbf{NTU RGB+D 120  (NTU-120)}~\cite{NTURGB120}: This dataset is the extension of NTU RGB+D 60 dataset by a number of performers and action classes, whose scale expands to 113,945 skeleton sequences covering 120 daily action classes. 
In detail, this dataset contains skeleton sequences of 106 different performers of a wide range of ages and it covers 155 camera views in 32 scenes. 
Two recommended evaluation metrics are used in this dataset: cross-subject (\textbf{Xsub}: skeleton sequences with 53 specific subject IDs are used for training and the remaining samples for testing) and cross-setup (\textbf{Xset}: skeleton sequences with even IDs are used for training while the remaining odd IDs for testing).  

\item \textbf{NTU RGB+D 61-120 (NTU-61-120)}~\cite{NTURGB120}: This dataset is the subset of NTU RGB+D 120, which contains 57,367 3D skeleton sequences covering the last 60 action classes in NTU RGB+D 120. 
The action classes in NTU RGB+D 61-120 have no intersection with the ones in NTU RGB+D 60. 
This dataset serves as an external dataset for evaluating the transfer capability of our DMMG.
\end{itemize}

\begin{table}[t]
    \caption{Linear evaluation results comparison on NTU-120 dataset.  \emph{Single-stream} denotes the best results represented by using the single data stream, \emph{Multi-stream} represents the best results represented by using multiple data streams.}
    \centering
    \scriptsize
    \begin{tabular}{l|c|cc}
        \toprule
        Method                    & Backbone        & Xsub (\%)       & Xset (\%)        \\      \midrule
        \multicolumn{4}{c}{\emph{Single-stream}}  \\ 
        LongT GAN~\cite{longtgan} &GRU          & 35.6	& 39.7  \\
        P\&C~\cite{pandc}         &RNN          & 42.7         & 41.7           \\
        CRRL~\cite{crrl}        &GRU         & 56.2	& 57.0           \\
        PCRP~\cite{PCRP}          &RNN       & 41.7          & 45.1           \\
        AS-CAL~\cite{AS-CAL}      &LSTM          & 48.6         & 49.2           \\
        ISC~\cite{isc}          &GCN          & 67.9	         & 67.1           \\
        SkeleMixCLR+~\cite{skeletonmix}  &GCN         & 69.0	         & 68.2            \\
        \textbf{DMMG (Ours)}       &GCN         & \textbf{69.6}	         & \textbf{70.1}            \\
        \hline
        \multicolumn{4}{c}{\emph{Multi-stream}}  \\ 
        3s-CrosSCLR (LSTM)~\cite{crossviews}        &LSTM             & 53.9          & 53.2           \\
        3s-SkeletonCLR~\cite{crossviews}     &GCN             & 60.7          & 62.6           \\
        3s-CrosSCLR~\cite{crossviews}        &GCN             & 67.9          & 66.7           \\
        3s-SkeleMixCLR+~\cite{skeletonmix}   &GCN              & 70.5          & 70.7          \\ 
        \textbf{2s-DMMG (Ours)}    &GCN     & \textbf{72.7}          &  \textbf{72.4}         \\ \bottomrule
    \end{tabular}
    \label{tab:ntu120-le}
\end{table}

\subsection{Evaluation Protocols}
\begin{table}[t]
    \caption{Comparison results of different self-supervised action recognition methods on the NTU-60 dataset, evaluated under the semi-supervised protocol.}
    \centering
    \scriptsize
    \tabcolsep1.4mm
    \begin{tabular}{l|c|ccc}
        \toprule
        \multirow{2}{*}{Method}  & \multirow{2}{*}{Label Fraction} & \multicolumn{2}{c}{NTU-60 (\%)} \\  
        & & Xsub  & Xview  \\ \midrule
        LongT GAN~\cite{longtgan}           & 1\%    & 33.1          & -             \\
        MS$^2$L~\cite{ms2l}                 & 1\%    & 35.2          & -             \\
        ISC~\cite{isc}                      & 1\%    & 35.7          & 38.1          \\
        3s-CrosSCLR~\cite{crossviews}       & 1\%    & 51.1          & 50.0          \\
        3s-Colorization~\cite{colorization} & 1\%    & 48.3          & 52.5          \\
        3s-AimCLR~\cite{extremeaug}         & 1\%   & 54.8          & 54.3           \\
        3s-SkeleMixCLR~\cite{skeletonmix}   & 1\%   & 55.3          & 55.7           \\
        \textbf{2s-DMMG (Ours)}             & 1\%   &\textbf{56.1}  &\textbf{56.6}   \\\midrule

        LongT GAN~\cite{longtgan}           & 10\%   & 62.0          & -             \\
        MS$^2$L~\cite{ms2l}                 & 10\%  & 65.2          & -            \\
        ISC~\cite{isc}                      & 10\%  & 65.9          & 72.5           \\
        3s-CrosSCLR~\cite{crossviews}       & 10\%  & 74.4          & 77.8          \\
        3s-Colorization~\cite{colorization} & 10\%  & 71.7          & 78.9         \\
        3s-AimCLR~\cite{extremeaug}         & 10\%  & 78.2          & 81.6           \\
        3s-SkeleMixCLR~\cite{skeletonmix}   & 10\%  & 79.9          & 83.6           \\ 
        \textbf{2s-DMMG (Ours)}             & 10\%  &\textbf{81.8} &\textbf{85.1}  \\\bottomrule
    \end{tabular}
    \label{semieval}
\end{table}

We evaluate our DMMG under four evaluation protocols, including linear, finetune, KNN, and semi-supervised evaluation protocols. 
\begin{itemize}
\item \textbf{Linear Evaluation Protocol.}
This protocol appends a fully connected layer with a Softmax activation function after a frozen pre-trained model and then uses a fully supervised method to train the classifier. 
We train the model for 80 epochs using an Adam optimizer with an initial learning rate of 0.001, which will be multiplied by 0.1 at epoch 60.
\item \textbf{Finetune Protocol.}
This protocol appends a linear classifier after a pre-trained model. 
Different from the linear evaluation protocol, the pre-trained model is trainable. We train the whole model using supervised learning. 
\item \textbf{KNN Evaluation Protocol.}
This protocol utilizes a k-nearest neighbor (KNN) classifier without trainable parameters to evaluate the quality of the action features encoded by the model.
We set K = 20, and the temperature parameter is 0.1 in the KNN evaluation protocol.
\item \textbf{Semi-Supervised Evaluation Protocol.}
In this protocol, we train the whole model under the finetune protocol by only using 1\% and 10\% randomly sampled labeled data, respectively.
\end{itemize}

\subsection{Comparison with State-of-the-Art Methods}
\subsubsection{Comparisons with Benchmark CLRs} 
We conduct our experiments on NTU-60 and NTU-120 datasets to compare our DMMG with four benchmarking contrastive learning-based self-supervised action recognition methods (e.g., SkeletonCLR~\cite{crossviews}, CrosSCLR~\cite{crossviews}, AimCLR~\cite{extremeaug}, and SkeleMixCLR~\cite{skeletonmix}).
Noted that the listed methods in Table~\ref{tab:linearevaul} utilize ST-GCN~\cite{stgcn} as the backbone, and DMMG$\ddagger$ denotes the model is pre-trained on NTU-61-120. 
SkeletonCLR is regarded as the baseline of the contrastive learning-based self-supervised action recognition method. 
It augments the skeleton sequences by using \emph{Shear} and \emph{Crop}.
SkeletonCLR considers the augmented skeleton sample as a positive sample and any other sample as a negative sample~\cite{crossviews}. 

As shown in Table~\ref{tab:linearevaul}, our DMMG achieves the best results on various streams, and our DMMG$\ddagger$ also achieves promising results. 
In addition, our DMMG further boosts the model's performance by only using the motion stream. 
This is because the motion representations are highly dependent on viewpoint variations. 
Our viewpoint variation min-max game augments the skeleton data with diverse motion information, and thus learns more representative skeleton motion features. 
Compared with three stream fusion methods, 3s-AimCLR~\cite{extremeaug} and 3s-SkeleMixCLR~\cite{skeletonmix}, our two stream fusion based DMMG (2s-DMMG) achieves higher accuracy by only using joint and motion streams. 
The experimental results demonstrate the effectiveness of our dual min-max games, which substantially improve the model's performance on action recognition task. 

\begin{figure*}[t]
    \centering
    \includegraphics[width=0.85\linewidth]{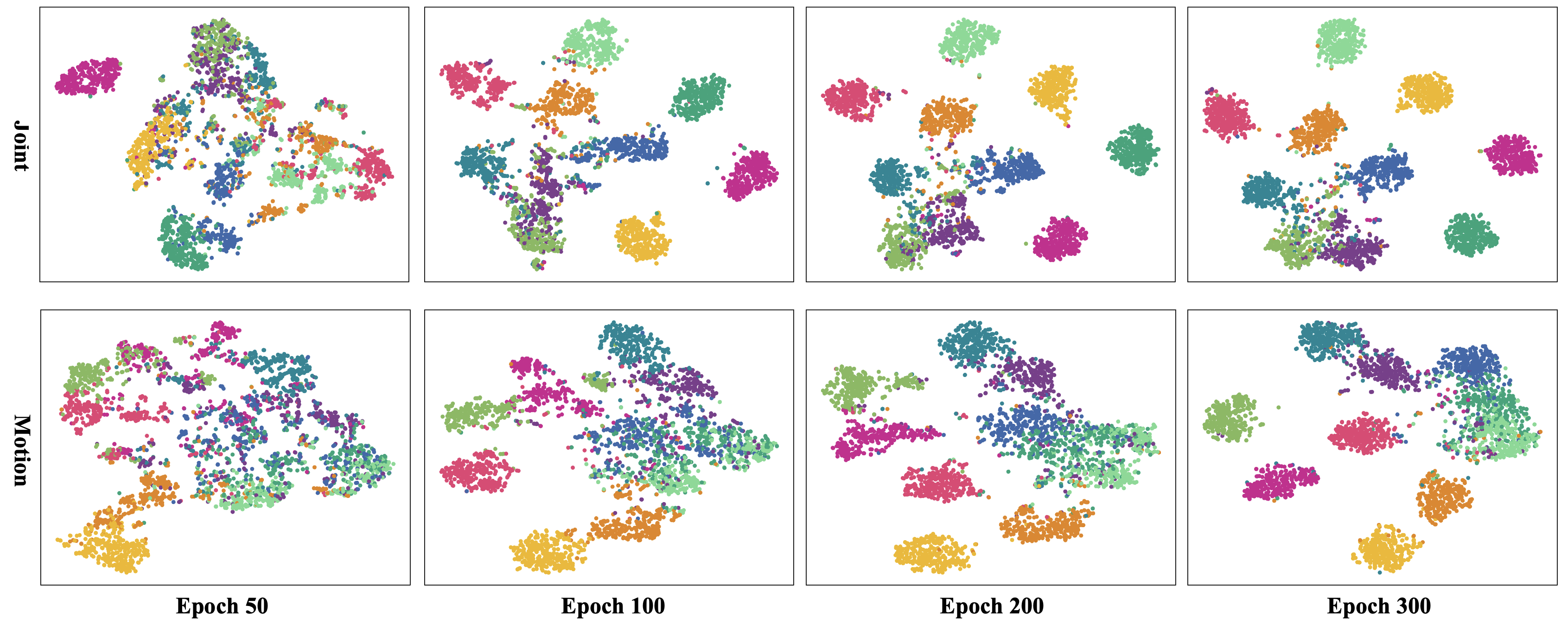}
    \caption{The t-SNE visualization of learned action features at different epochs during pre-training on NTU-60 Xview.}
    \label{fig:qualitative}
\end{figure*}

\subsubsection{Linear Evaluation Protocol Results} 
Table~\ref{tab:ntu60-le} and Table~\ref{tab:ntu120-le} demonstrate the results compared with other state-of-the-art methods under linear evaluation protocol on NTU-60 and NTU-120 datasets, respectively. 
Here, we compare our method with the approaches that use various deep models as backbones.

From the single-stream perspective, it can be observed that there is a significant gap in accuracy between GCN-based methods (e.g., ISC~\cite{isc}) and other model-based methods (e.g., LSTM-based AS-CAL~\cite{AS-CAL}, GRU-based CRRL~\cite{crrl}).
Compared with the GCN-based methods, our DMMG achieves the best results. 
From the multi-stream perspective, our 2s-DMMG further improves the performance and outperforms many other multi-stream methods, such as 3s-AimCLR~\cite{extremeaug}, 3s-CrosSCLR~\cite{crossviews}, 3s-SkeletonCLR~\cite{crossviews}, and 3s-Colorization~\cite{colorization}, on both NTU-60 and NTU-120 datasets. 
The superior performance of our 2s-DMMG on small-scale and large-scale datasets (NTU-60 and NTU-120) demonstrates the effectiveness and generalization of our dual min-max games. 

\begin{table}[t]
    \caption{Results under the finetune protocol on NTU-60 and NTU-120 datasets. ``$\S$'' indicates the model is trained with fully-supervised learning and $\ddagger$ indicates the model is pre-trained on NTU-61-120. The results of 2s-ST-GCN$^\S$ are reproduced on our settings.}
    \centering
    \scriptsize
    \begin{tabular}{l|cccc}
        \toprule
        \multirow{2}{*}{Method} & \multicolumn{2}{c}{NTU-60 (\%)} & \multicolumn{2}{c}{NTU-120 (\%)} \\
        & Xsub           & Xview           & Xsub            & Xset            \\ \midrule
        2s-ST-GCN$^\S$~\cite{stgcn}      & 85.0           & 91.2            & 77.0            & 77.2            \\
        3s-CrosSCLR$\ddagger$~\cite{crossviews}             & 85.6           & 92.0           & -            & -            \\
        3s-CrosSCLR~\cite{crossviews}             & 86.2           & 92.5            & 80.5            & 80.4            \\
        3s-AimCLR~\cite{extremeaug}             & 86.9           & 92.8             & 80.1            & 80.9            \\
        3s-SkeleMixCLR~\cite{skeletonmix}       & 87.7  & 93.9     & 81.6   & 81.2 \\
        2s-DMMG$\ddagger$ (Ours)      &86.8	 &93.7	& -	&- \\
        \textbf{2s-DMMG (Ours)}      & \textbf{87.9}  & \textbf{94.2}      & \textbf{82.4}   & \textbf{83.0}\\ \bottomrule
    \end{tabular}
    \label{tab:finetune}
\end{table}

\subsubsection{Finetune Protocol Results}
Table~\ref{tab:finetune} shows the results comparison of our methods with other benchmark methods under the finetune protocol. 
Noted that the ST-GCN is trained with fully supervision.
As shown in Table~\ref{tab:finetune}, our 2s-DMMG achieves the best results. 
Compared to the results under the linear evaluation protocol, the results under finetune protocol show a significant improvement. 
Our 2s-DMMG performs slightly better than 3s-SkeleMixCLR under the finetune protocol. 
This is because utilizing various data streams can contribute more to model performance improvement, but using more data streams comes with a higher computational cost. 
Our dual min-max games can boost performance with a relatively lower computational load.

\subsubsection{KNN Evaluation Protocol Results}
\begin{table}[t]
    \caption{Comparisons of our DMMG with SkeletonCLR, AimCLR, and SkeleMixCLR under the KNN evaluation protocol by using the joint stream.}
    \centering
    \scriptsize
    \tabcolsep0.9mm
    \begin{tabular}{l|cccc}
        \toprule
        \multirow{2}{*}{Method} & \multicolumn{2}{c}{NTU-60 (\%)} & \multicolumn{2}{c}{NTU-120 (\%)}\\
        & Xsub           & Xview           & Xsub            & Xset        \\ \midrule
        SkeletonCLR~\cite{crossviews}  & 64.8  & 60.7 & 41.9 & 42.9 \\
        AimCLR~\cite{extremeaug}  & 71.0 & 63.7 & 48.9 & 47.3 \\
        SkeleMixCLR~\cite{skeletonmix}   & 72.3  & 65.5 & 49.3 & 48.3 \\ 
        \textbf{DMMG (Ours)}  & \textbf{72.8}  & \textbf{69.6} & \textbf{51.5} & \textbf{52.3} \\
        \bottomrule
    \end{tabular}
    \label{knneval}
\end{table}

As shown in Table~\ref{knneval}, our DMMG achieves better results than SkeletonCLR, AimCLR, and SkeleMixCLR on both datasets. 
Notably, our DMMG achieves a significant improvement in action recognition accuracy on Xview metrics in the NTU-60 dataset. 
The results under the KNN evaluation protocol demonstrate that our viewpoint-variant game can guide the model to learn more discriminative feature representations from various viewpoints.

\subsubsection{Semi-Supervised Protocol Results} 
To ensure each action class has roughly equal representation in the training samples, we randomly sample 1\% and 10\% labeled data from each action class, respectively. 
As shown in Table~\ref{semieval}, our 2s-DMMG outperforms other methods. 
This indicates that our dual min-max games can make full use of spatio-temporal information to enable the model to perform better on the downstream task.

\begin{figure}[t]
    \centering
    \includegraphics[width=1\linewidth]{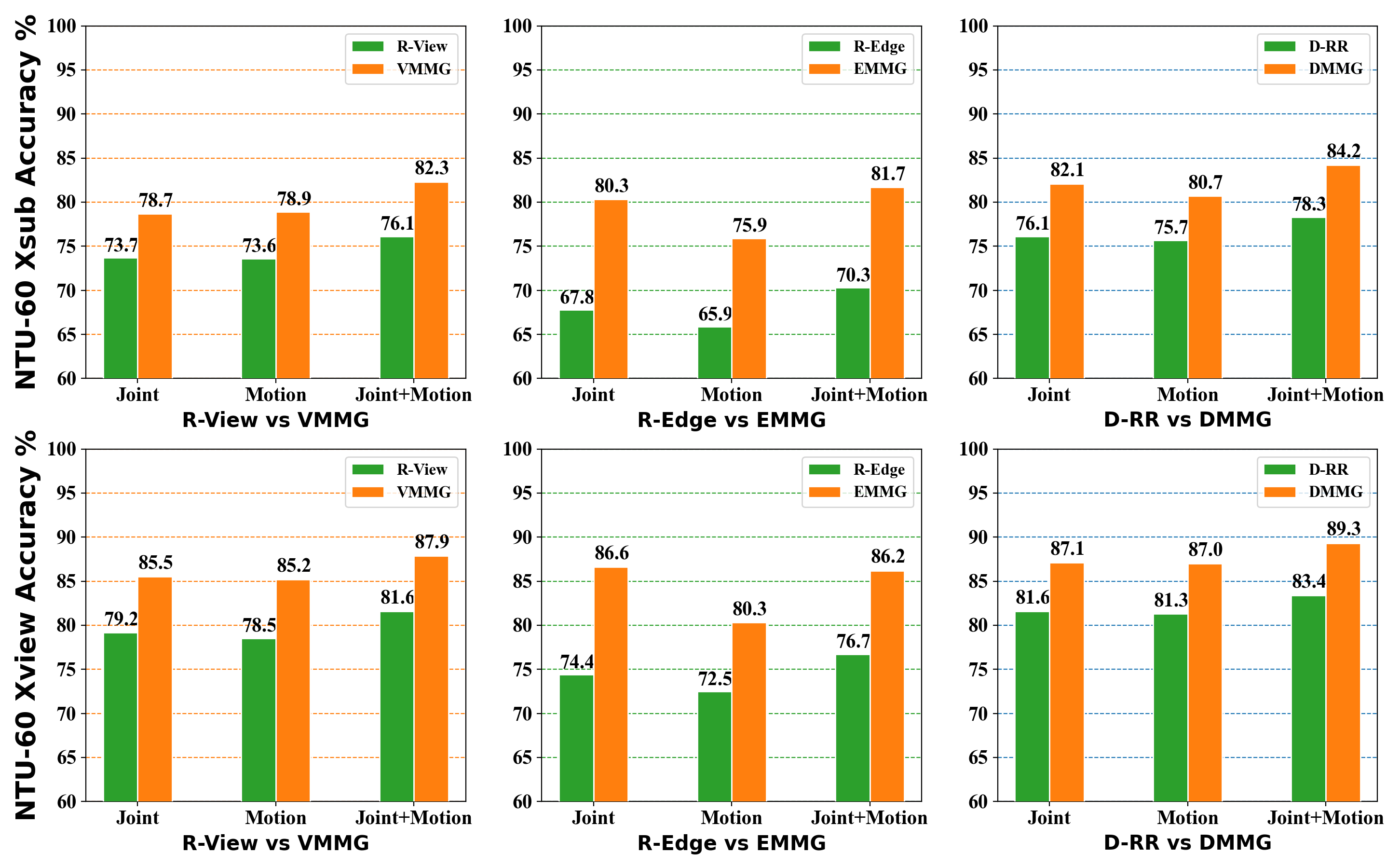}
    \caption{Linear evaluation results to evaluate the effectiveness of the min-max game strategy. VMMG denotes the model trained with the viewpoint variant min-max game, and EMMG represents the model trained by the edge perturbation min-max game. R-View and R-Edge denote the random data augmentation strategy from the viewpoint and edge perturbation perspectives, respectively, while D-RR denotes the combination of VMMG and EMMG.}
    \label{fig:absrandmmg}
\end{figure}

\begin{figure*}[t]
    \centering
    \includegraphics[width=0.85\linewidth]{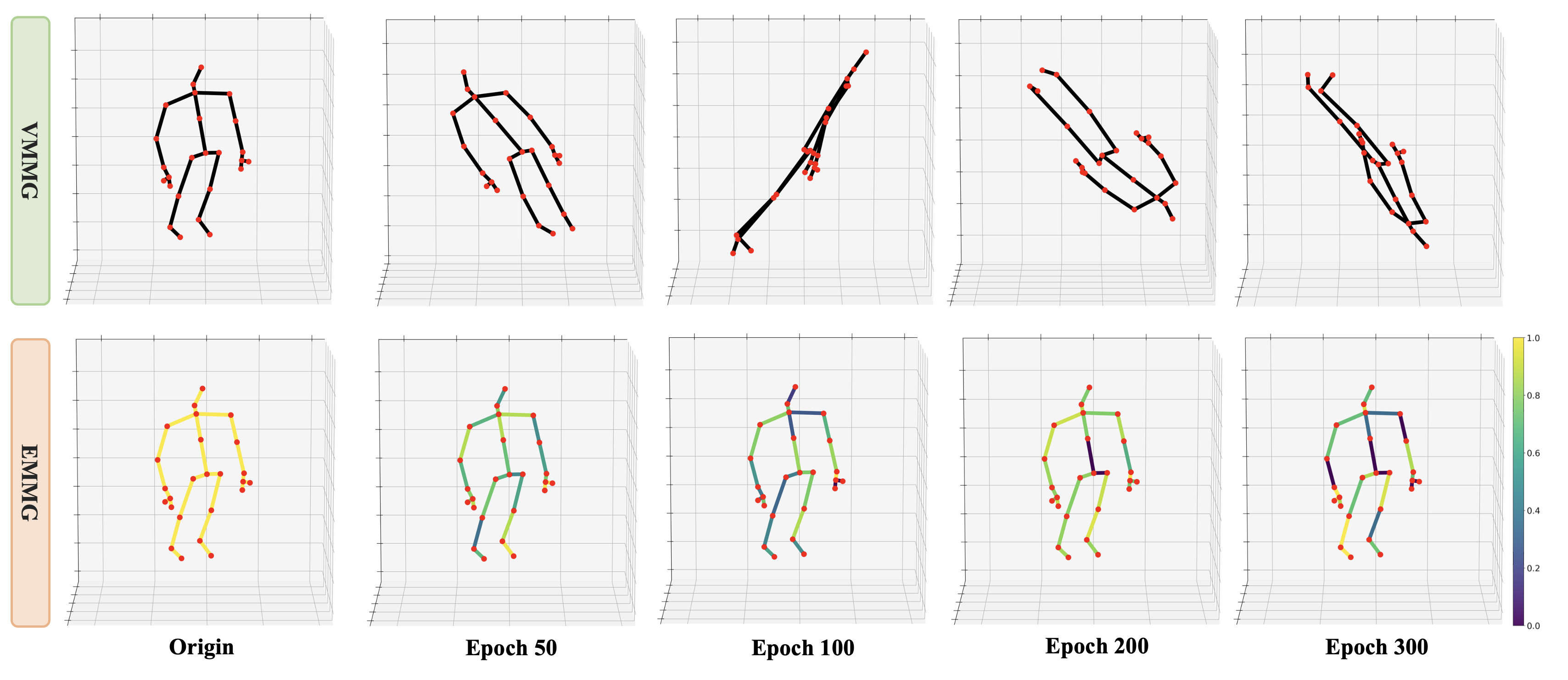}
    \caption{Visualization of augmented data by DMMG on sitting-down action samples of NTU-60. Results of VMMG show the skeleton data augmented by the viewpoint variant min-max game, and results of EMMG represents the graph-structured body joints augmented by the edge perturbation min-max game. In the results of EMMG, the yellow color indicates a stronger connection between the two joints.}
    \label{fig:visualizeofdmmg}
\end{figure*}

\subsubsection{Qualitative Results}
We use t-SNE~\cite{tsne} to visualize the action features learned by the pre-trained model at 50, 100, 200, 300 epochs. 
For a fair comparison, we randomly select ten action categories for feature visualization.
The feature visualization results in Fig~\ref{fig:qualitative} show that our DMMG makes the action feature representations of the same category more clustered and distinguishable as training proceeds. 
The feature representations of the motion stream are discriminative, leading to the promising results in Table~\ref{tab:linearevaul}.
The qualitative results demonstrate that our DMMG can learn more discriminative features and thus boost the model's performance on downstream tasks.

\begin{figure}[t]
    \centering
    \includegraphics[width=0.95\linewidth]{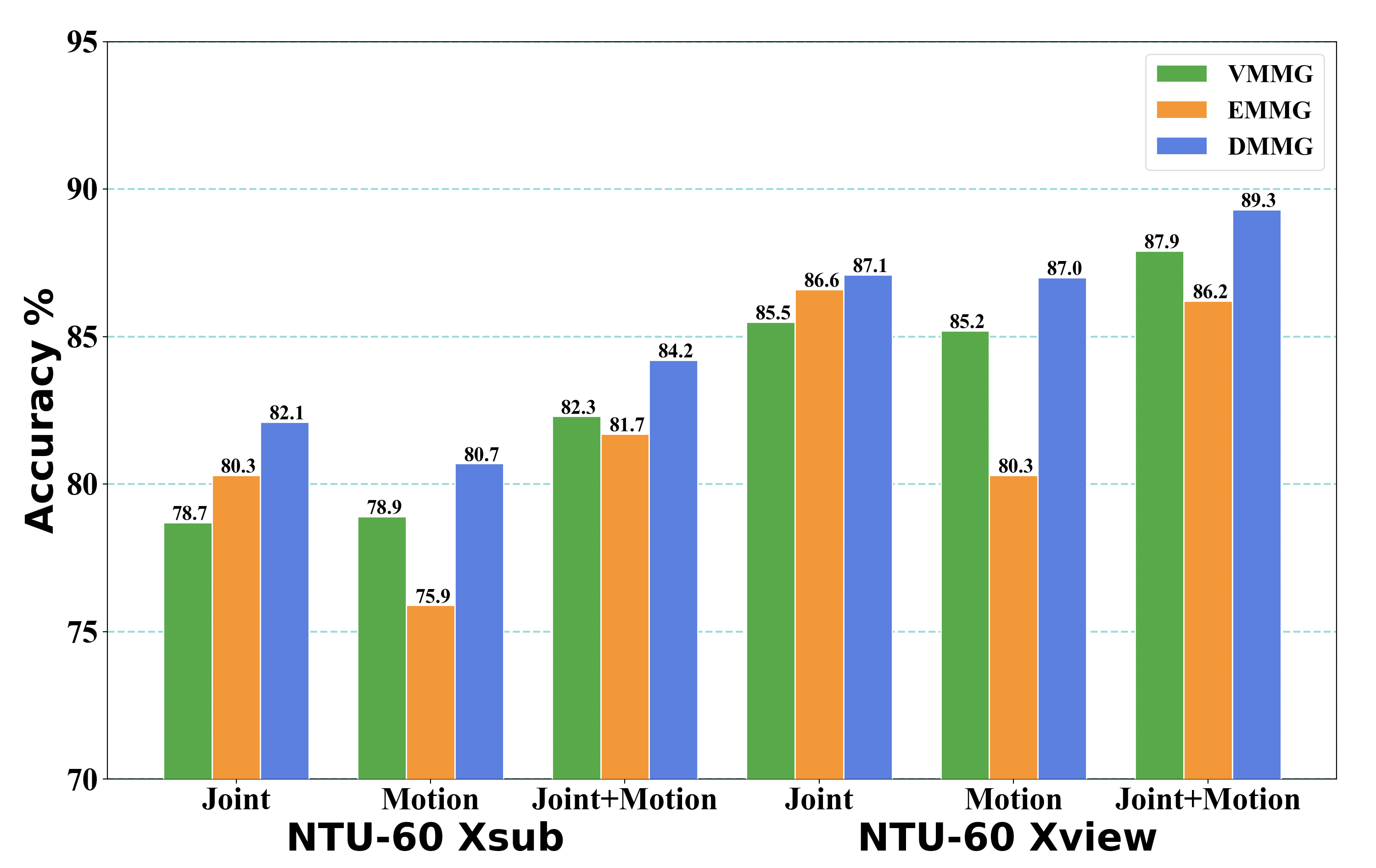}
    \caption{Comparison of Linear evaluation results to show the effectiveness of Dual Min-Max games.}
    \label{fig:absinglemmg}
\end{figure}

\subsection{Ablation Study}
We conduct five ablation studies to verify the effectiveness of different components of our DMMG. 
Here, we denote the viewpoint variant min-max game as VMMG, and use EMMG to represent the edge perturbation min-max game.

\subsubsection{Effectiveness of the Min-Max Game Strategy}
To evaluate the effectiveness of our min-max game strategy, we design two random data augmentation strategies to substitute the two min-max games. 
Firstly, we randomly change the viewpoints to augment the skeleton sequence. 
We denote such a random viewpoint variation augmentation strategy as R-View.
Secondly, we randomly perturb the connectivity strengths among graph-based body joints to augment graph-based body joints, and denote this random edge perturbation augmentation strategy as R-Edge.  
Finally, we use D-RR to represent the combination of R-View and R-Edge.

From Fig~\ref{fig:absrandmmg}, we can observe that DMMG outperforms D-RR, and the min-max game strategies (VMMG and EMMG) have a better performance than both R-View and R-Edge. 
This verifies the adversarial paradigm enables our DMMG to construct more challenging contrastive pairs. 
In addition, we can find that the performance of R-Edge is worse than R-View. 
This is because a purely random edge perturbation strategy may drop connections between key joints, which may cause the model cannot capture critical action information. 

\subsubsection{Effectiveness of Dual Min-Max Games}
To evaluate the effectiveness of our two min-max games, we train the model using only one min-max game. 
For a fair comparison, we adopt the same data streams, training settings and evaluation protocol as in Table~\ref{tab:linearevaul}, and we conduct the experiments on the NTU-60 dataset.
As shown in Figure~\ref{fig:absinglemmg}, we can observe that VMMG achieves higher accuracy than EMMG when using the motion stream, while EMMG outperforms VMMG when only using the joint stream. 
This verifies that our two min-max games can improve the model's performance from different data streams, and combining the two min-max games can further improve the model's performance.

\subsubsection{Transfer Ability} 
To evaluate the transfer ability of our DMMG, we first conduct experiments on the NTU-61-120 dataset, then transfer the pre-trained model on NTU-60 for linear and finetune evaluation. 
As shown in Table~\ref{tab:linearevaul}, we can find that our DMMG$\ddagger$ achieves better results than many other methods on the NTU-60 dataset. 
Especially under Xview protocol, DMMG$\ddagger$ has comparable performance with DMMG. 
In Table~\ref{tab:finetune}, our DMMG$\ddagger$ achieves better performance than 3s-CrosSCLR$\ddagger$~\cite{crossviews}.
The experimental results demonstrate the strong transfer ability of our DMMG.

\subsubsection{Comparisons of Different GCN Models} 
To evaluate the effectiveness of our DMMG on different GCN models, we conduct experiments by using different GCN models on NTU-60 and NTU-120 datasets. 
Here, we employ 2s-ASGCN~\cite{asgcn} as the candidate GCN model. 
As one of the most robust GCN models, 2s-ASGCN achieves state-of-the-art performance on both NTU-60 and NTU-120 datasets. 

As shown in Table~\ref{tab:evaluationonasgcn}, our 2s-DMMG with the ASGCN model achieves better results than 2s-DMMG using the STGCN model under both Linear and Finetune evaluation protocols. 
This means our DMMG can achieve better performance on downstream tasks when using a more robust GCN model. 
Compared with 2s-ASGCN under a fully-supervised training manner, our 2s-DMMG has slightly improved performance under the finetune protocol on NTU-60. 
This is because the 2s-ASGCN can learn sufficiently discriminative action features and achieve excellent performance on NTU-60. 
As a result, our pre-trained model can only achieve limited improvement.

\begin{table}[t]
    \caption{Comparison of Results with Different GCN Models on NTU-60 and NTU-120 datasets.}
    \centering
    \scriptsize
    \begin{tabular}{l|c|cccc}
        \toprule
        \multirow{2}{*}{Method} &\multirow{2}{*}{Evaluation} & \multicolumn{2}{c}{NTU-60 (\%)} & \multicolumn{2}{c}{NTU-120 (\%)} \\
        &  & Xsub           & Xview           & Xsub            & Xset  \\ \midrule
        2s-STGCN~\cite{stgcn} &  Fully-Supervised  & 85.0 &	91.2 & 77.0	&77.2          \\
        2s-ASGCN~\cite{asgcn} & Fully-Supervised  & 88.5	& 95.1	& 80.5	& 82.6           \\
        2s-DMMG (STGCN) & Linear  & 84.2	& 89.3	& 72.7	& 72.4           \\
        2s-DMMG (ASGCN) & Linear  & 86.1	& 92.7	& 76.2	& 78.9     \\
        2s-DMMG (STGCN) & Finetune  & 87.9	& 94.2	& 82.4	& 83.0 \\
        2s-DMMG (ASGCN) & Finetune & \textbf{88.7} & \textbf{95.2} & \textbf{83.3} & \textbf{84.2} \\ \bottomrule
    \end{tabular}
    \label{tab:evaluationonasgcn}
\end{table}

\subsubsection{Visualization of DMMG}
We visualize the original data and augmented data from epochs 50, 100, 200, and 300. 
As shown in Fig~\ref{fig:visualizeofdmmg}, the difference between the augmented data and original data increases with training epochs, and the viewpoint gradually becomes more challenging in the viewpoint variant min-max game.
As indicated by the results of EMMG, we can observe that more connectivity strengths among graph-based body joints decrease with increasing training epochs. 

Especially, in epochs 200 to 300, some key joint connection values are close to zero, indicating that some unnecessary connections are ignored.
The visualization of augmented data by our DMMG demonstrates the effectiveness of the min-max game strategy in producing challenging contrastive pairs. 
These contrastive pairs can help the model learn more discriminative action feature representations, thus improving its performance on downstream tasks.

\section{Conclusion}
In this paper, we proposed Dual Min-Max Games (DMMG) based data augmentation for a self-supervised skeleton-based action recognition task under a contrastive learning framework. 
Our DMMG has a viewpoint variation min-max game and an edge perturbation min-max game. 
They perform augmentation on the skeleton sequences and graph-structured body joints, respectively.
As a result, we obtain sufficient challenging contrastive pairs with our DMMG.
These hard contrastive pairs help the model to learn more representative and discriminative feature representations while avoiding overfitting. Thus, our method significantly boosts the performance of our model on downstream tasks.
The extensive experimental results demonstrate the effectiveness of our DMMG and show that our DMMG achieves state-of-the-art performance under various evaluation protocols.

\bibliographystyle{IEEEtran}
\bibliography{mybib}
%



\end{document}